%% file: main.tex
\documentclass[10pt,twocolumn,letterpaper]{article}

\usepackage[pagenumbers]{cvpr} 

\input{preamble}
\usepackage{colortbl}
\usepackage{xcolor}
\usepackage{booktabs}

\definecolor{myblue}{rgb}{0.8, 0.9, 1.0}    
\definecolor{mygreen}{rgb}{0.9, 1.0, 0.9}   
\definecolor{myorange}{rgb}{1.0, 0.9, 0.8}  
\definecolor{myred}{rgb}{1.0, 0.85, 0.85}   

\definecolor{mycream}{HTML}{FFFEE0}
\definecolor{cvprblue}{rgb}{0.21,0.49,0.74}
\usepackage[pagebackref,breaklinks,colorlinks,allcolors=cvprblue]{hyperref}

\title{Pest-Thinker: Learning to Think and Reason like Entomologists via Reinforcement Learning}

\author{
Xueheng Li$^{1,2}$\thanks{Equal contribution.}
\quad
Yu Wang$^{1,2}$\footnotemark[1]
\quad
Tao Hu$^{1,2}$
\quad
Ji Huang$^{1,2}$
\quad
Ke Cao$^{1,2}$
\\
Qize Yang$^{4}$
\quad
Rui Li$^{1}$
\quad
Jie Zhang$^{1,3}$
\quad
Chengjun Xie$^{1,3}$\thanks{Corresponding author.}
\\
$^{1}$Institute of Intelligent Machines, Hefei Institute of Physical Science, Chinese Academy of Sciences
\\
$^{2}$University of Science and Technology of China
\\
$^{3}$Zhongke Hefei Institute of Technology Innovation Engineering
\\
$^{4}$Hefei University of Technology
\\
{\tt\small lixueheng@mail.ustc.edu.cn, \{lirui, zhangjie, cjxie\}@iim.ac.cn}
}

\begin{document}
 \maketitle

\input{sec/0_abstract}    
\vspace{-0.5em}
\input{sec/1_intro}
\input{sec/2_relatedwork}

\input{sec/4_Method}

\input{sec/5_Experiment}

\input{sec/6_Conclusion}
{
    \small
    \bibliographystyle{ieeenat_fullname}
    \bibliography{main}
}


\end{document}


\maketitle

\section{Detailed Prompts Used in the Project}
\label{sec:prompts}

This section presents all the prompt templates used in our project. The prompts are organized into three main categories: (1) CoT generation for dataset creation, (2) feature reward evaluation using LLM judge, and (3) model answer generation during training.

\subsection{CoT Generation Prompts}

As illustrated in the table-style layout of Figure~\ref{fig:prompt_cot_system}, the following prompts are used to generate high-quality Chain-of-Thought (CoT) reasoning data for the QFSD-CoT-2K and AgriInsect-CoT-3K datasets. These prompts guide the model to produce natural, expert-like reasoning processes for agricultural pest identification.

\begin{figure*}[t]
    \centering
    \begin{tcolorbox}[
        title=\textbf{Prompt Template for CoT Generation (System Prompt)},
        colback=gray!10,
        colframe=gray!75!black,
        boxrule=0.5pt,
        left=1em, right=1em,
        top=0.5em, bottom=1em,
        width=\textwidth, 
        fontupper=\small
        ]
        You are an expert entomologist identifying agricultural pests from field images.

        When analyzing the image, think naturally like a human expert examining an insect. Include what's most striking or noticeable about the insect, its color, shape, size, or distinctive markings. Describe what you see using natural language, compare the observable features with each option provided, and explain your reasoning process as you narrow down to the correct identification. Please refer to the provided knowledge, but do not reveal that you are using existing knowledge.

        \textbf{Guidelines:}
        \begin{itemize}[leftmargin=*,nosep,itemsep=0.1em]
            \item Begin your analysis by describing the most noticeable or distinctive features you observe.
            \item Express your thoughts naturally as you examine different parts of the insect.
            \item Consider each option systematically and explain why features match or don't match.
            \item Think through your identification process like explaining your reasoning to a colleague.
        \end{itemize}
    \end{tcolorbox}
    \caption{System prompt for CoT data generation.}
    \label{fig:prompt_cot_system}
\end{figure*}

The user prompt template shown in Figure~\ref{fig:prompt_cot_user} provides the specific question, answer options, and reference knowledge about the target pest species. The placeholders (e.g., \texttt{\{options\_str\}}, \texttt{\{core\_feature\_info\}}) are dynamically filled with actual values during dataset generation.

\begin{figure*}[t]
    \centering
    \begin{tcolorbox}[
        title=\textbf{Prompt Template for CoT Generation (User Prompt)},
        colback=gray!10,
        colframe=gray!75!black,
        boxrule=0.5pt,
        left=1em, right=1em,
        top=0.5em, bottom=1em,
        width=\textwidth, 
        fontupper=\small
        ]
        Which of the following pests is shown in the image?

        \textbf{Options:}\\
        \texttt{\{options\_str\}}

        \textbf{Reference knowledge about} \texttt{(\{latin\_name\}):}\\
        \texttt{Category: \{major\_category\}}\\
        \texttt{\{core\_feature\_info\}}

        Analyze the image carefully and think through your identification naturally. Describe the observable features, compare them with each option, and explain your reasoning process.

        \textbf{Format:}\\
        \texttt{<think>}\\
        \texttt{[Your natural reasoning process - describe what you observe, evaluate each option, and reach your conclusion]}\\
        \texttt{</think>}\\
        \texttt{<answer>\{correct\_letter\}</answer>}
    \end{tcolorbox}
    \caption{User prompt for CoT data generation.}
    \label{fig:prompt_cot_user}
\end{figure*}

\subsection{Feature Reward (LLM Judge) Prompts}

The feature reward mechanism uses an LLM judge to evaluate how well the model's reasoning process covers the ground truth morphological features, as shown in Figure~\ref{fig:prompt_reward_system}. The following prompts are used to instruct the LLM judge to score the quality of the model's reasoning.

\begin{figure*}[t]
    \centering
\begin{tcolorbox}[
	title=\textbf{Prompt Template for Feature Reward (LLM Judge System Prompt)},
	title after break=\textbf{Prompt Template for Feature Reward (LLM Judge System Prompt) — Continued},
	colback=gray!10,
	colframe=gray!75!black,
	breakable,
	boxrule=0.5pt,
	left=1em, right=1em,
	top=0.5em, bottom=1em,
	width=\textwidth
	]
	You are an expert entomology judge. Your task is to evaluate the quality of an AI assistant's reasoning process for identifying an insect based on a set of ground truth features.

	You will be provided with:
	\begin{enumerate}[leftmargin=*,nosep,itemsep=0.1em]
		\item Ground Truth Features: A list of key morphological features of the insect.
		\item Model Reasoning: The internal monologue (\texttt{<think>} block) generated by the AI assistant, where it describes the features it observes from an image.
	\end{enumerate}

	Your goal is to score how well the model's reasoning covers and correctly describes the provided ground truth features.

	\textbf{Scoring Rubric (0-10 scale):}
	\begin{itemize}[leftmargin=*,nosep,itemsep=0.1em]
		\item \textbf{10 (Excellent):} The reasoning explicitly and accurately describes almost all available ground truth features.
		\item \textbf{7-9 (Good):} The reasoning correctly identifies a majority of the key ground truth features.
		\item \textbf{5-6 (Moderate):} The reasoning mentions some correct features but misses several important ones, or contains minor inaccuracies.
		\item \textbf{3-4 (Poor):} The reasoning misses most of the ground truth features or contains significant inaccuracies.
		\item \textbf{0-2 (Failure):} The reasoning is completely irrelevant, hallucinatory, or fails to describe any correct features.
	\end{itemize}

	\textbf{Scoring Guideline:}
	\begin{enumerate}[leftmargin=*,nosep,itemsep=0.1em]
		\item The model reasoning process may include language descriptions such as horizontal comparisons of various answer options. You need to read the model reasoning process carefully to find out whether the model's description of key insect characteristics is accurate and comprehensive.
		\item Think like a professional human expert and provide a comprehensive and credible rating.
		\item Encourage reflection and further examination of the model's reasoning content.
	\end{enumerate}

	\textbf{CRITICAL OUTPUT FORMATTING RULES:}\\
	You MUST provide only a single score. Your entire response MUST be in the format \texttt{Score: X}, where X is a number between 0 and 10. Do NOT add any other text, explanation, or justification.

	\textbf{Example of a perfect response:}\\
	\texttt{Score: 6.5}
\end{tcolorbox}
\caption{System prompt for the LLM-as-a-Judge.}
\label{fig:prompt_reward_system}
\end{figure*}


As illustrated in Figure~\ref{fig:prompt_reward_user}, the user prompt template for the LLM judge includes the ground truth features and the model's reasoning content. The judge evaluates the reasoning and returns a score between 0 and 10, which is then normalized to a reward value between 0.0 and 1.0.

\begin{figure*}[t]
    \centering
\begin{tcolorbox}[
	title=\textbf{Prompt Template for Feature Reward (LLM Judge User Prompt)},
	title after break=\textbf{Prompt Template for Feature Reward (LLM Judge User Prompt) — Continued},
	colback=gray!10,
	colframe=gray!75!black,
	breakable,
	boxrule=0.5pt,
	left=1em, right=1em,
	top=0.5em, bottom=1em,
	width=\textwidth
	]
	\textbf{Ground Truth Features:}\\
	\texttt{\{formatted\_features\}}

	\textbf{Model's Reasoning (\texttt{<think>} block):}\\
	\texttt{\{model\_think\_content\}}

	---\\
	Based on the rubric, please provide a score for the model's reasoning. Remember, your response must be ONLY in the format \texttt{Score: X}.
\end{tcolorbox}
\caption{User prompt for the LLM-as-a-Judge.}
\label{fig:prompt_reward_user}
\end{figure*}

\subsection{Model Answer Generation Prompts}

During GRPO training, the prompts shown in Figure~\ref{fig:prompt_answer_system} guide the model to generate answers with detailed reasoning processes. The system prompt establishes the model's role and output format, while the user prompt template provides the specific question and instructions for the reasoning process.

\begin{figure*}[t]
    \centering
\begin{tcolorbox}[
	title=\textbf{Prompt Template for Model Answer Generation (System Prompt)},
	title after break=\textbf{Prompt Template for Model Answer Generation (System Prompt) — Continued},
	colback=gray!10,
	colframe=gray!75!black,
	breakable,
	boxrule=0.5pt,
	left=1em, right=1em,
	top=0.5em, bottom=1em,
	width=\textwidth
	]
	You are an expert assistant in entomology and agriculture. The user will provide an image and ask a question. The assistant first thinks about the reasoning process in the mind, meticulously observing and describing the insect's morphological features (like head, thorax, abdomen, wings, legs, body surface and special markings.), and then provides the user with the answer. The reasoning process and answer are enclosed within \texttt{<think>} \texttt{</think>} and \texttt{<answer>} \texttt{</answer>} tags, respectively, i.e., \texttt{<think>} reasoning process here \texttt{</think>}\texttt{<answer>} answer here \texttt{</answer>}
\end{tcolorbox}
\caption{System prompt for the Model Answer Generation.}
\label{fig:prompt_answer_system}
\end{figure*}

The user prompt template in Figure~\ref{fig:prompt_answer_user} encourages the model to engage in natural, human-like reasoning with internal dialogue and self-reflection. The model is required to provide detailed morphological observations and reasoning before giving the final answer.

\vspace{0.5em}
\begin{figure*}[t]
    \centering
\begin{tcolorbox}[
	title=\textbf{Prompt Template for Model Answer Generation (User Prompt)},
	title after break=\textbf{Prompt Template for Model Answer Generation (User Prompt) — Continued},
	colback=gray!10,
	colframe=gray!75!black,
	breakable,
	boxrule=0.5pt,
	left=1em, right=1em,
	top=0.5em, bottom=1em,
	width=\textwidth
	]
	\texttt{\{Question\}}

	Please act as an agricultural expert and think about this question as if you were a human pondering deeply. Engage in an internal dialogue, observing the morphological details, patterns, and colors. Use expressions such as `let me think...', `let's break it down...', `I notice that...', `this characteristic suggests...', or other natural thought expressions. It's encouraged to include self-reflection or verification in your reasoning process. Provide your detailed reasoning between the \texttt{<think>} \texttt{</think>} tags, and then give your final answer between the \texttt{<answer>} \texttt{</answer>} tags. Please provide only the single option letter (e.g., A, B, C, D, etc.) within the \texttt{<answer>} \texttt{</answer>} tags.
\end{tcolorbox}
\caption{User prompt for the Model Answer Generation.}
\label{fig:prompt_answer_user}
\end{figure*}

\section{Knowledge Base Format}

This section, illustrated in Figures~\ref{fig:kb_format} and~\ref{fig:kb_example}, presents the format of the knowledge base used in our project. The knowledge base is stored in JSON format and contains morphological feature descriptions for agricultural pest species. Each entry includes taxonomic information and detailed core features organized by body parts.

\subsection{Knowledge Base Entry Format}

The knowledge base file contains an array of pest entries. Each entry follows the structure shown in Figure~\ref{fig:kb_format}, where \texttt{core\_features} is a nested object containing feature descriptions for different morphological aspects of the insect.

\begin{figure*}[t]
    \centering
\begin{tcolorbox}[
	title=\textbf{Knowledge Base Entry Format},
	title after break=\textbf{Knowledge Base Entry Format — Continued},
	colback=gray!10,
	colframe=gray!75!black,
	breakable,
	boxrule=0.5pt,
	left=1em, right=1em,
	top=0.5em, bottom=1em,
	width=\textwidth
	]
\begin{verbatim}
{
  "category_id": ...,
  "latin_name": ...,
  "major_category": ...,
  "core_features": {
    "Overall Body Size and Shape": ...,
    "Head Features (Shape, Color, Eyes, Antennae, etc.)": ...,
    "Thorax Features (Pronotum, Scutellum, etc.)": ...,
    "Abdomen Features (Shape, Color, Segments, etc.)": ...,
    "Wing Features (Shape, Color, Patterns, Transparency, etc.)": ...,
    "Leg Features (Color, Morphology, Special Structures, etc.)": ...,
    "Body Surface Features (Hair, Luster, Texture, etc.)": ...,
    "Special Markings (Spots, Stripes, Special Structures, etc.)": ...
  }
}
\end{verbatim}
\end{tcolorbox}
\caption{Structured knowledge base entry format.}
\label{fig:kb_format}
\end{figure*}

This knowledge base format enables structured access to morphological features during dataset generation and training. The nested structure allows for easy retrieval of specific feature categories, which are used in the CoT generation prompts and feature reward evaluation.

\subsection{Example Knowledge Base Entry}

The following example, shown in Figure~\ref{fig:kb_example}, demonstrates a complete knowledge base entry with actual feature descriptions.

\begin{figure*}[t]
    \centering
\begin{tcolorbox}[
	title=\textbf{Example Knowledge Base Entry},
	title after break=\textbf{Example Knowledge Base Entry — Continued},
	colback=gray!10,
	colframe=gray!75!black,
	breakable,
	boxrule=0.5pt,
	left=1em, right=1em,
	top=0.5em, bottom=1em,
	width=\textwidth
	]
\{
\vspace{0.2em}\\
\phantom{XX}\texttt{"category\_id"}: 20000,\\
\phantom{XX}\texttt{"latin\_name"}: Abidama liuensis Metcalf \& Horton,\\
\phantom{XX}\texttt{"major\_category"}: cicada,\\
\phantom{XX}\texttt{"core\_features"}:\\
\phantom{XX}\{\\
\phantom{XXXX}\texttt{"Overall Body Size and Shape"}: Body length 8-9mm.,\\
\phantom{XXXX}\texttt{"Head Features (Shape, Color, Eyes, Antennae, etc.)"}: Head black, shiny. Male head conical and protruding forward, female head without this feature. Compound eyes purplish brown.,\\
\phantom{XXXX}\texttt{"Thorax Features (Pronotum, Scutellum, etc.)"}: Thorax black, shiny. Scutellum triangular, raised in the middle.,\\
\phantom{XXXX}\texttt{"Abdomen Features (Shape, Color, Segments, etc.)"}: Abdomen deep yellow.,\\
\phantom{XXXX}\texttt{"Wing Features (Shape, Color, Patterns, Transparency, etc.)"}: Forewings blackish brown, distinctly depressed below the claw suture. Two orange-yellow spots near the base. Two orange-yellow club-shaped spots near the apex (one large, one small, the small spot difficult to see clearly).,\\
\phantom{XXXX}\texttt{"Leg Features (Color, Morphology, Special Structures, etc.)"}: Tibiae and tarsi of thoracic legs blackish brown.,\\
\phantom{XXXX}\texttt{"Body Surface Features (Hair, Luster, Texture, etc.)"}: Head and thorax shiny.,\\
\phantom{XXXX}\texttt{"Special Markings (Spots, Stripes, Special Structures, etc.)"}: Orange-yellow spots near the base and apex of the forewings. Male head conical and protruding forward.\\
\phantom{XX}\}\\
\}
\end{tcolorbox}
\caption{Example of a populated knowledge base entry.}
\label{fig:kb_example}
\end{figure*}

\section{Details of Baseline Model}

\begin{figure}[t]
	\centering
        \includegraphics[width=0.48\textwidth]{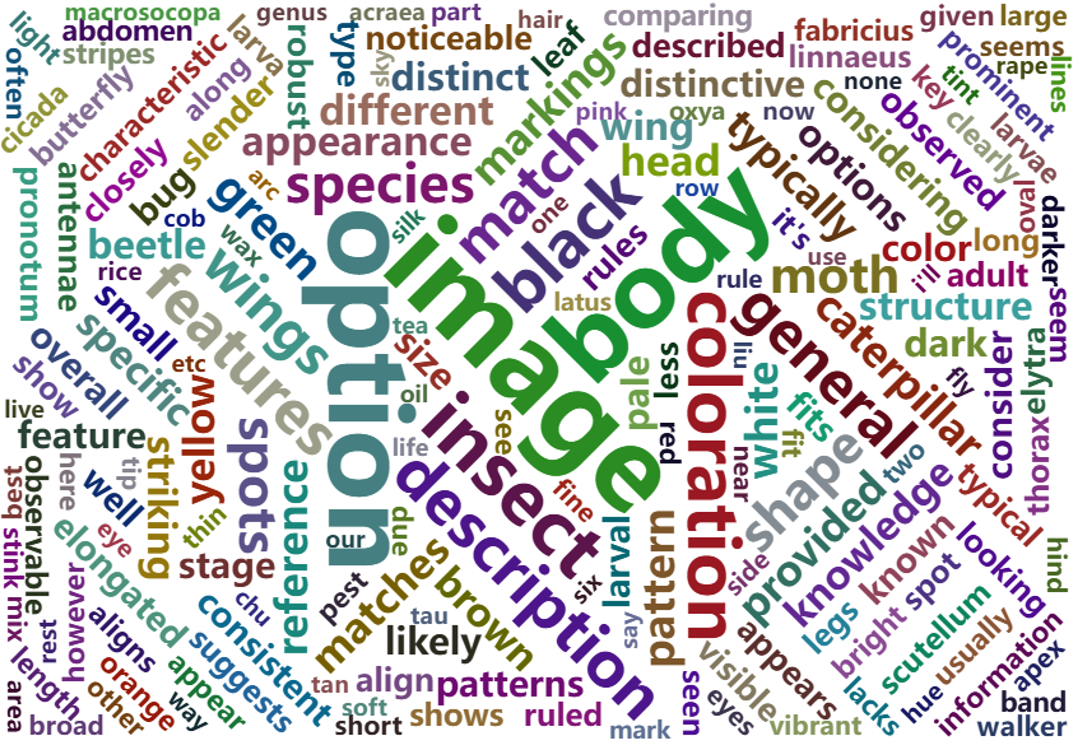}
	\caption{Word cloud built from the cold-start CoT corpora (QFSD-CoT-2K and AgriInsect-CoT-3K), presenting high-frequency morphological descriptors.}
\label{fig:supp_wordcloud}
\end{figure}

\noindent \textbf{GPT-5} \citet{openai2025gpt5} marks a substantial advance in multimodal reasoning and large-scale world modeling. According to OpenAI's announcement and technical overview, GPT-5 shows notable gains on academic and human-evaluated benchmarks (including math, coding, and multimodal tasks) and is presented as a general-purpose multimodal reasoning system that integrates stronger visual understanding and higher task performance compared with prior releases.

\noindent \textbf{GPT-5-mini} \cite{openai2025gpt5} is a more compact variant of GPT-5, designed for efficient deployment while maintaining strong multimodal reasoning capabilities. It offers a balance between performance and computational efficiency, making it suitable for applications requiring faster inference and lower resource consumption.

\noindent \textbf{Gemini-2.5-Flash} \cite{comanici2025gemini} is described by DeepMind/Google as a "workhorse" thinking model optimized for fast, low-cost inference suitable for everyday tasks such as summarization, chat, and captioning. Public product and model pages emphasize its design tradeoffs for speed and value at scale and list it among the Gemini-2.5 family intended to support low-latency multimodal applications.

\noindent \textbf{InternVL3-8B} \cite{zhu2025internvl3} represents a new milestone in multimodal large language models (MLLMs). It adopts a \textit{native multimodal pre-training} strategy, jointly training on large-scale text and vision-language data rather than sequentially fine-tuning, which enables the model to align linguistic and visual representations from the ground up. The model further introduces the \textit{Variable Visual Position Encoding} (V2PE) mechanism to handle complex spatial layouts and longer visual contexts. With these architectural refinements, InternVL3-8B achieves state-of-the-art results on a wide range of multimodal understanding benchmarks, including document OCR, chart reasoning, and fine-grained visual perception tasks.

\noindent \textbf{InternVL3.5-8B} \cite{wang2025internvl3} extends the InternVL3 family with substantial improvements in reasoning efficiency and deployment flexibility. It integrates a two-stage \textit{Cascade Reinforcement Learning} (Cascade RL) pipeline—combining offline and online reinforcement learning—to enhance multimodal reasoning capability. Furthermore, a \textit{Visual Resolution Router} (ViR) dynamically adjusts visual token resolution to optimize speed-performance trade-offs, and the \textit{Decoupled Vision-Language Deployment} (DvD) framework allows distributed computation across heterogeneous GPUs. Experimental evaluations demonstrate that InternVL3.5-8B improves average multimodal benchmark scores by over 16\% compared with its predecessor, while achieving up to 4$\times$ inference acceleration. Beyond traditional perception and understanding tasks, InternVL3.5 also supports embodied interaction and GUI reasoning, expanding its applicability to intelligent agent scenarios.

\noindent \textbf{Qwen2.5VL-Instruct series} \cite{bai2025qwen2} is presented as a major iteration in the Qwen vision-language line: the Qwen2.5-VL family improves visual recognition, document parsing, and long-video comprehension, and the released materials document a variety of instruction-tuned and vision-language variants intended for practical multimodal tasks. Research on reinforcement-style fine-tuning and guided thought methods has also been evaluated on Qwen2.5 agents in the literature.

\noindent \textbf{Qwen3VL-Instruct series} \cite{yang2025qwen3} extends the Qwen visual-language family toward larger, reasoning-enhanced variants. Public model repositories and model cards describe Qwen3-VL instruction-tuned variants that emphasize deeper visual perception, temporal reasoning, and multi-stage decoding for cross-modal tasks.

\noindent \textbf{Qwen3VL-Thinking series} \cite{yang2025qwen3} represents the "thinking" editions of the Qwen3-VL line, designed to enhance reasoning capabilities through explicit thought processes. These variants incorporate multi-stage reasoning mechanisms and structured thinking patterns to improve performance on complex visual reasoning tasks.

\noindent \textbf{VisionThink-Efficient-7B} \cite{yang2025visionthink} is a variant of VisionThink optimized for efficiency, framing visual reasoning efficiency as a learned decision process (for example, reducing visual tokens while preserving VQA performance). The VisionThink project reports improvements on general VQA benchmarks using reinforcement-style training and an LLM-as-judge strategy to balance efficiency with reasoning quality.

\noindent \textbf{VisionThink-General-7B} \cite{yang2025visionthink} is a more general variant of VisionThink that emphasizes comprehensive visual reasoning capabilities. It demonstrates strong performance across diverse visual question answering tasks while maintaining efficient inference characteristics through learned decision processes.

\noindent \textbf{Vision-SR1-7B} \cite{li2025self} explores self-rewarding or stepwise reflective reinforcement approaches to improve visual reasoning without substantial external supervision. This work decomposes VLM reasoning into perception and language stages and demonstrates that RL-style objectives can refine intermediate perceptual outputs and downstream reasoning. The Vision-SR1 preprints and codebases summarize empirical gains and methodologies in this vein.

\noindent \textbf{Vision-G1-7B} \cite{zha2025vision} is framed as an attempt to scale general vision-language reasoning through multi-round RL training and guided reinforcement procedures (GRR). The paper outlines constructing RL-ready datasets spanning many tasks and using curriculum/data-selection strategies with reinforcement objectives to improve generalization across diverse visual tasks.

\section{Benchmark Visualization}
\label{sec:benchmark_visual}

Our benchmark datasets exhibit rich visual diversity and inherent task difficulty. AgriInsect contains 9{,}452 high-resolution field photographs spanning 200 species, while QFSD comprises 7{,}054 images across 141 species captured under occlusion and illumination shifts. Together they expose models to a broad spectrum of Lepidoptera, Hemiptera, Coleoptera, and other economically important orders. To further increase the fine-grained challenge, every multiple-choice question draws all distractor options from the same major taxonomic category as the ground-truth species, forcing models to closely inspect fine-grained visual cues rather than relying on coarse cues.

\begin{figure*}[!htbp]
	\centering
        \includegraphics[width=\linewidth]{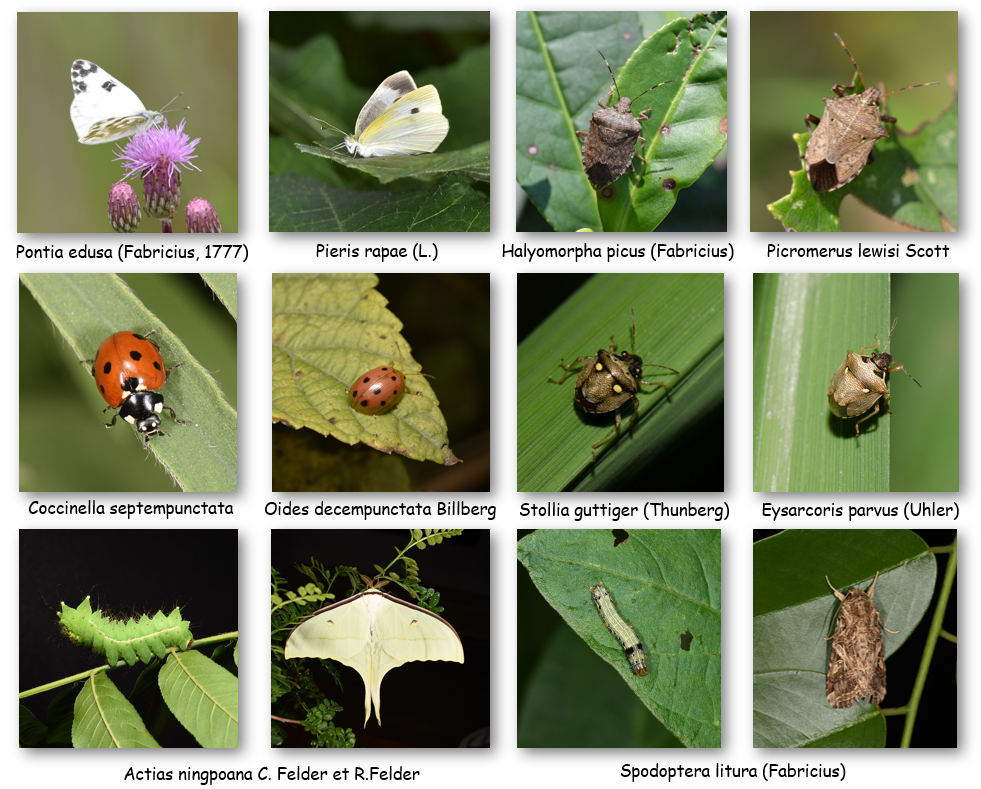}
	\caption{Representative agricultural pests from the AgriInsect and QFSD benchmarks, illustrating the fine-grained visual diversity considered in our study.}
\label{fig:supp_case}
\end{figure*}

Figure~\ref{fig:supp_case} juxtaposes representative pest images with their scientific names. The first two rows deliberately mix distinct taxonomic categories with morphologically similar appearances, underscoring how visual similarity alone can be misleading, while the bottom row emphasizes intra-species diversity by showing markedly different developmental stages. Figure~\ref{fig:supp_wordcloud} presents a word cloud derived from the cold-start CoT corpora (QFSD-CoT-2K and AgriInsect-CoT-3K), summarizing high-frequency descriptors such as \textit{elytra}, \textit{thorax}, \textit{hindwing}, and \textit{pronotum}, thereby illustrating the morphology-aware vocabulary encountered by the model during reasoning.

\section{RL Training Curve}
\begin{figure*}[!htbp]
	\centering
        \includegraphics[width=\linewidth]{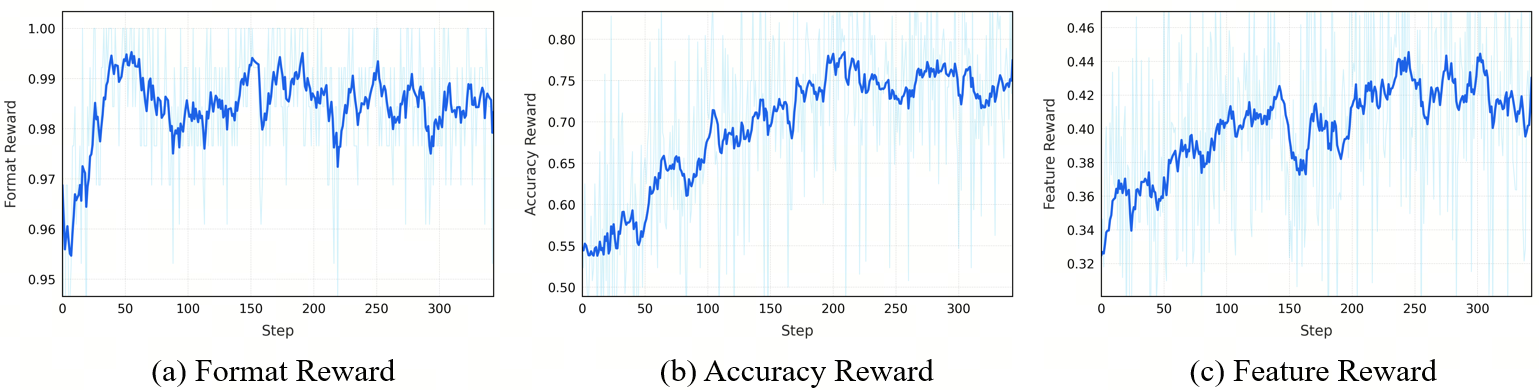}
	\caption{The reward function training curves on the AgriInsect dataset.}
\label{FIG:reward}
\end{figure*}

Figure \ref{FIG:reward} illustrates the reward training curves during the RL stage on the AgriInsect dataset across the full-set experiments. Consistently, the accuracy and feature rewards exhibit an upward trend, demonstrating that GRPO effectively drives progressive improvements in model performance.

\section{Case Study}
\begin{figure*}[!htbp]
	\centering
        \includegraphics[width=\linewidth]{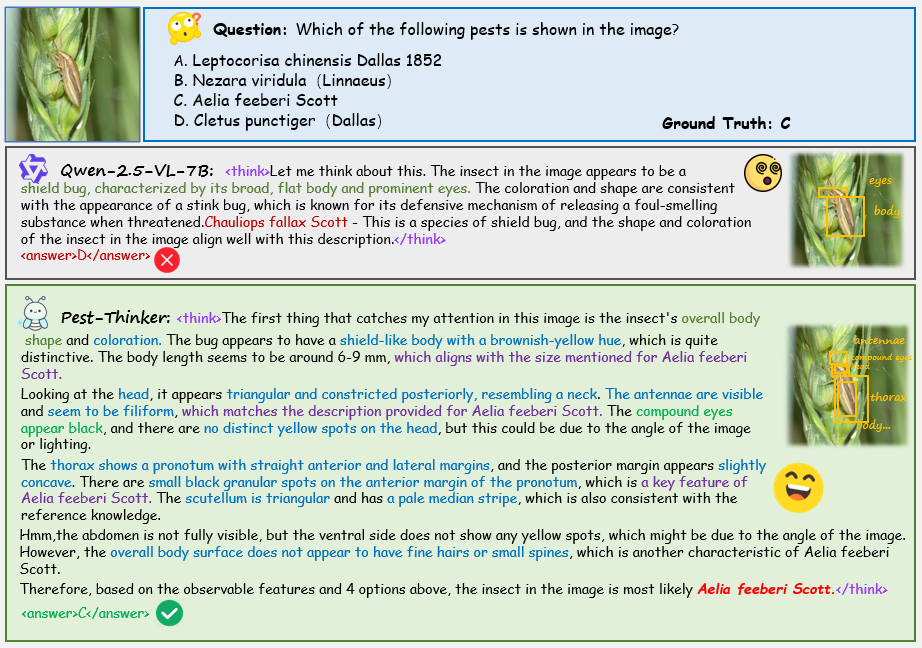}
	\caption{Comparison of reasoning traces generated by Qwen2.5-VL and Pest-Thinker on the QFSD dataset for Bug (\textit{Aelia feeberi Scott}).}
\label{FIG:case02}
\end{figure*}

This section presents detailed case studies comparing the reasoning processes of Qwen2.5-VL and Pest-Thinker on various pest identification tasks from the QFSD dataset. The following figures illustrate how Pest-Thinker's reinforcement learning-based training enables more comprehensive morphological feature analysis and more accurate identification compared to the baseline model. Each case demonstrates the model's ability to observe and reason about key insect characteristics, including body structure, coloration, wing patterns, and other distinguishing features.

\begin{figure*}[!htbp]
	\centering
        \includegraphics[width=\linewidth]{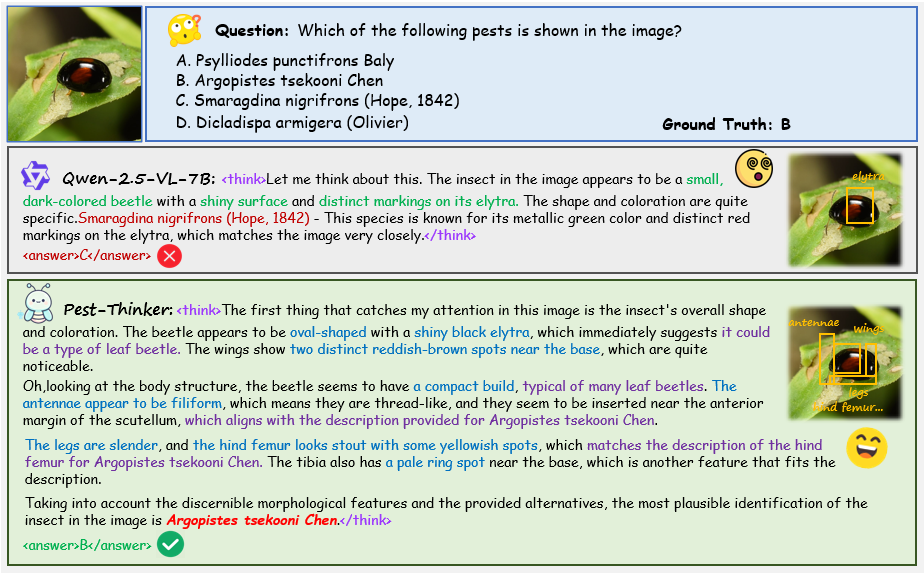}
	\caption{Comparison of reasoning traces generated by Qwen2.5-VL and Pest-Thinker on the QFSD dataset for Beetle (\textit{Argopistes tsekooni Chen}).}
\label{FIG:case03}
\end{figure*}

\begin{figure*}[!htbp]
	\centering
        \includegraphics[width=\linewidth]{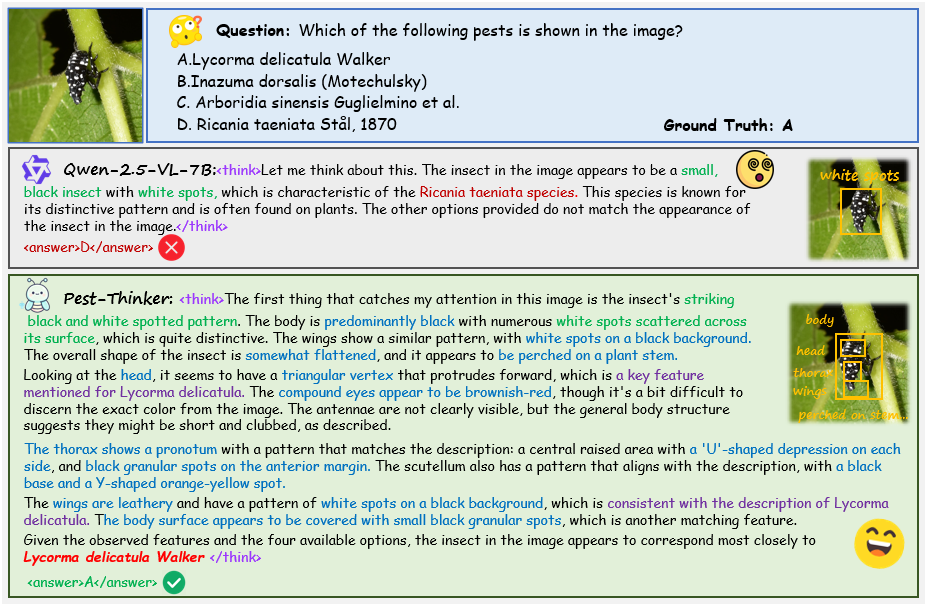}
	\caption{Comparison of reasoning traces generated by Qwen2.5-VL and Pest-Thinker on the AgriInsect dataset for Cicada (\textit{Lycorma delicatula Walker}).}
\label{FIG:case04}
\end{figure*}

\begin{figure*}[!htbp]
	\centering
        \includegraphics[width=\linewidth]{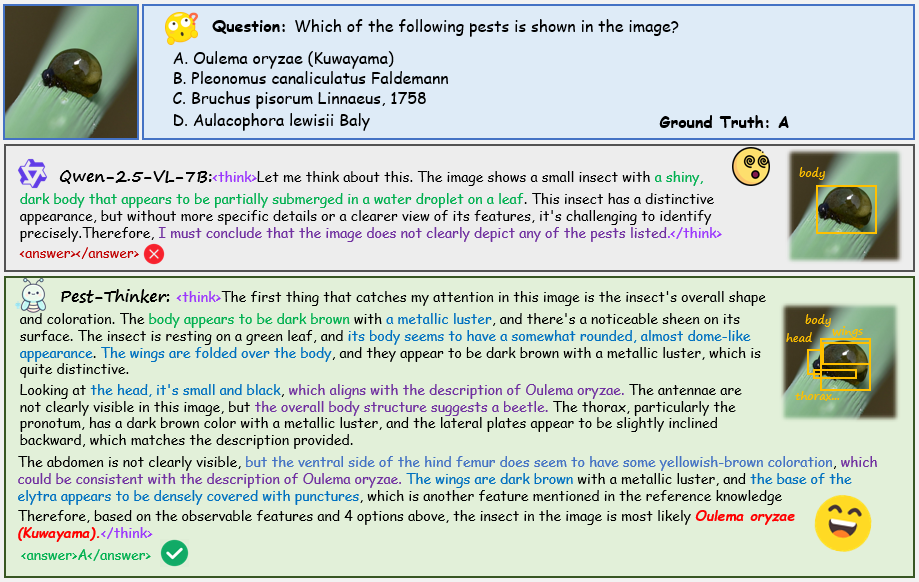}
	\caption{Comparison of reasoning traces generated by Qwen2.5-VL and Pest-Thinker on the AgriInsect dataset for Beetle (\textit{Oulema oryzae (Kuwayama)}).}
\label{FIG:case05}
\end{figure*}

\begin{figure*}[!htbp]
	\centering
        \includegraphics[width=\linewidth]{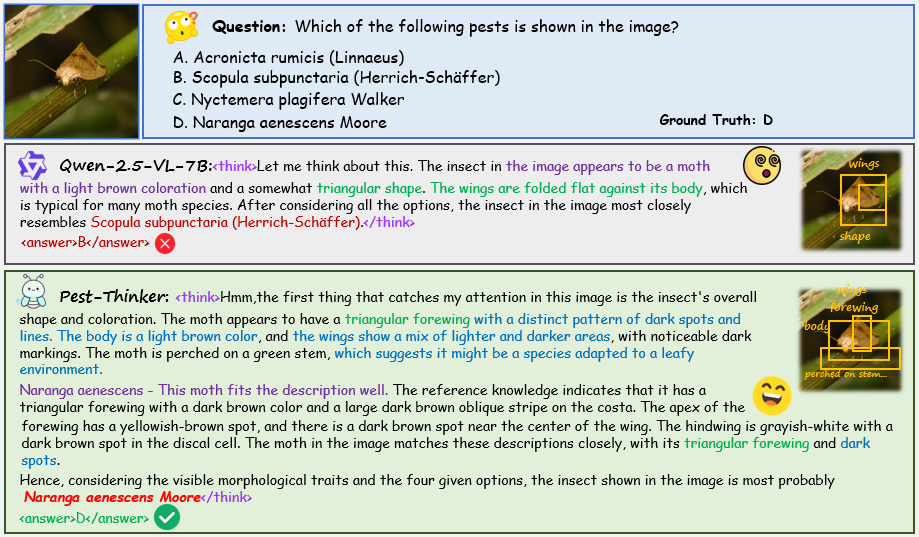}
	\caption{Comparison of reasoning traces generated by Qwen2.5-VL and Pest-Thinker on the AgriInsect dataset for Moth (\textit{Naranga aenescens Moore}).}
\label{FIG:case06}
\end{figure*}

Figure~\ref{FIG:case02} shows a comparison for Bug (\textit{Argopistes tsekooni} Chen), Figure~\ref{FIG:case03} presents the case for Beetle (\textit{Argopistes tsekooni} Chen), Figure~\ref{FIG:case04} illustrates the reasoning process for Cicada (\textit{Lycorma delicatula} Walker), Figure~\ref{FIG:case05} demonstrates the comparison for Beetle (\textit{Oulema oryzae} (Kuwayama)), and Figure~\ref{FIG:case06} displays the case for Moth (\textit{Naranga aenescens} Moore). These examples highlight Pest-Thinker's enhanced capability to systematically examine morphological features and provide detailed reasoning before making the final identification.

{
    \small
    \bibliographystyle{ieeenat_fullname}
    \bibliography{main}
}

%% file: preamble.tex
\usepackage{tcolorbox}
\usepackage{amssymb}  
\usepackage{xcolor}    
\usepackage{enumitem}  



%% file: sec/0_abstract.tex
\begin{abstract}
Pest-induced crop losses pose a major threat to global food security and sustainable agricultural development. While recent advances in Multimodal Large Language Models (MLLMs) have shown strong potential for visual understanding and smart agriculture, their direct application to pest recognition remains limited due to the domain’s unique challenges such as high inter-species complexity, intra-species variability, and the scarcity of expert-annotated data. In this work, we introduce Pest-Thinker, a knowledge-driven reinforcement learning (RL) framework that enables MLLMs to reason over fine-grained pest morphology. We first construct two high-definition pest benchmarks, QFSD and AgriInsect, comprising diverse species and expert-annotated morphological traits. Leveraging these datasets, we synthesize Chain-of-Thought (CoT) reasoning trajectories to facilitate structured learning of pest-specific visual cues through Supervised Fine-Tuning (SFT). Subsequently, we employ Group Relative Policy Optimization (GRPO) with a novel feature reward that guides the model to focus on observable morphological evidence, assessed by an LLM-as-a-Judge strategy. Extensive experiments demonstrate that Pest-Thinker substantially improves both in-domain and out-of-domain morphological understanding, marking a step toward expert-level visual reasoning for intelligent agricultural pest analysis. The datasets and source code are available upon acceptance.

\end{abstract}

%% file: sec/1_intro.tex
\section{Introduction}
\label{sec:intro}
\begin{figure}[t]
	\centering
        \includegraphics[width=\linewidth]{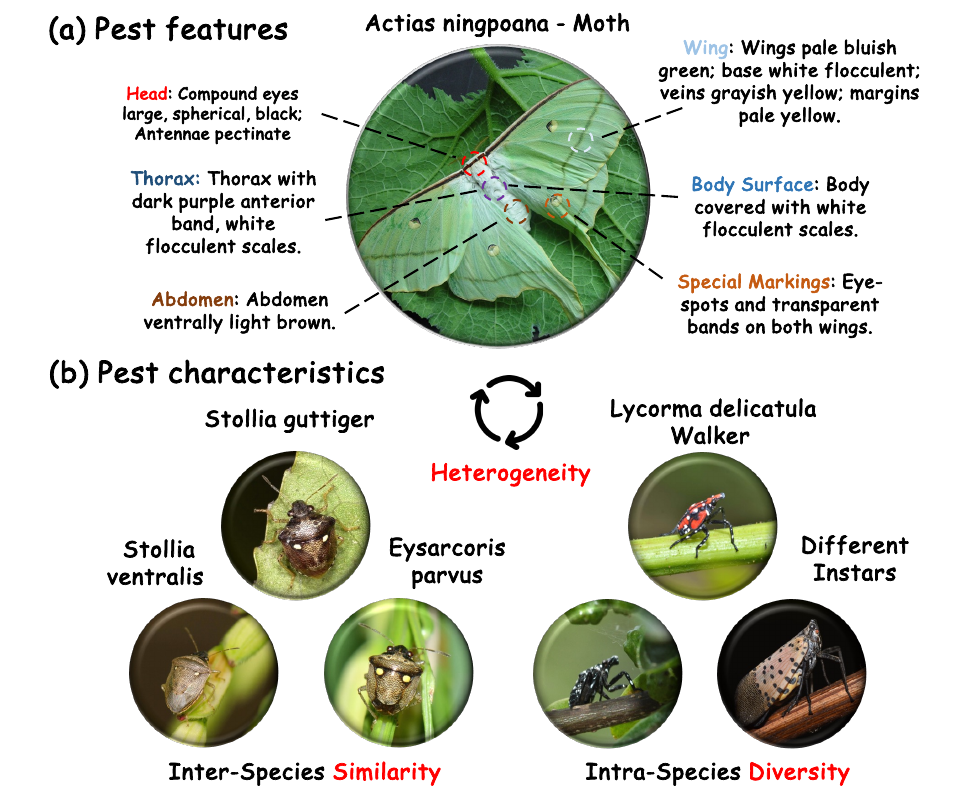}
	\caption{The complex morphological features of pests, coupled with their inherent heterogeneity, similarity, and diversity, pose significant challenges to the understanding and reasoning of pest-related visual information.}
\label{FIG:1}
\end{figure}
Pest-induced damage to crops remains a major threat to global food security \cite{karp2016agricultural}, economic stability, and sustainable agricultural development \cite{paini2016global}. On average, the cash crops of a single region may be affected by dozens of pest species, leading to substantial yield losses and ecological imbalance \cite{oerke2006crop, bruce2010tackling}. Consequently, the development of effective and intelligent pest management and recognition strategies is essential to ensure stable agricultural productivity and promote sustainable farming systems.

In recent years, Multimodal Large Language Models (MLLMs) \cite{comanici2025gemini, bai2025qwen2, zhu2025internvl3, hong2025glm, team2025kimi} have garnered widespread attention for their remarkable performance and have pioneered revolutionary paradigms in a wide range of visual understanding tasks. The rapid advancement of MLLMs has also stimulated increasing research interest in their applications within the field of smart agriculture. Representative works, such as Insect-Foundation \cite{truong2025insect} and AgriGPT-VL \cite{yang2025agrigptvl, yang2025agrigpt}, have introduced large-scale agriculture-oriented MLLMs through extensive domain-specific training dataset. However, existing approaches remain predominantly reliant on data-driven Supervised Fine-Tuning (SFT) on MLLMs. Inherently constrained by extensive data requirements, large model scales, proprietary access, and high computational demands, these methods are often ill-suited for the practical applications in real-world agricultural scenarios.

In contrast to general-purpose visual understanding, pest recognition and perception in agricultural contexts require models to accurately capture and learn the core morphological features of diverse pest species. However, the visual attributes of pests are often subtle, complex, and highly variable. As illustrated in Figure \ref{FIG:1}(a), critical morphological features (e.g., the eyes, thorax, wings, and distinctive body surface markings) exhibit substantial variation across species. Furthermore, as shown in Figure \ref{FIG:1}(b), not only do closely related pest species (e.g., \textit{Stollia guttiger} vs. \textit{Stollia ventralis}) exhibit nearly identical visual characteristics, but even individuals of the same species can display notable different appearances (e.g., at various instars and growth stages) \cite{munir2019crop}. This combination of high inter-species complexity and intra-species appearance diversity poses a fundamental challenge for general MLLMs to accurately perceive and reason about pest-specific visual features. Moreover, existing pest benchmarks often suffer from low image resolution, coarse-grained species categorization, and a lack of expert-level morphological annotations. Effective expertise is essential for building robust post-training benchmarks and promoting a better understanding of MLLMs. These factors substantially hinder effective knowledge transfer and limit the practical adaptability of MLLMs for real-world agricultural applications.

Recently, the ``\textit{Thinking with images}” paradigm \cite{openai2025thinking, zheng2025deepeyes, su2025thinking} based on Reinforcement Learning (RL) \cite{guo2025deepseek, shen2025vlm} have significantly enhanced the reasoning and problem-solving capabilities of MLLMs in understanding and interpreting visual information. This dynamic approach to multimodal reasoning has yielded notable progress across diverse visual reasoning tasks, including visual question answering \cite{zhan2025vision, liu2025visionreasoner} and complex scene understanding \cite{su2025pixel, deng2025openvlthinker}. RL-based post-training methods for MLLMs, compared to SFT, are more data-efficient and foster superior generalizability by facilitating learning from trial and mistakes \cite{liu2025visual, tan2025reason}. This advantage renders RL methods particularly well-suited for agricultural applications, where high-quality labeled data is often scarce. However, although existing RL methods have substantially enhanced the reasoning and cognitive abilities of MLLMs, their applications in the agricultural domain remain largely underexplored. More critically, the training strategies adopted in prior work are insufficient to equip MLLMs to capture and reason about the underlying causal relationships (i.e., morphology and physical characteristics) during training. Merely transferring generic visual RL frameworks cannot cultivate the specialized reasoning required for intricate pest understanding tasks, particularly the ability to generalize reasoning processes across  multiple pest species with diverse visual features.


To address the aforementioned challenges, we build two high-definition pest benchmarks, QFSD and AgriInsect, encompassing a wide range of pest species with clear morphological features. Compared with existing pest datasets, our data provides fine-grained classifications (e.g., \textit{Cicada} vs. \textit{Abidama liuensis}), more accurately reflecting the real-world data distribution. To facilitate the learning of pest visual features, we further compile expert-annotated pest feature knowledge provided by agricultural specialists and categorized the core morphological traits according to distinct body parts (as shown in Figure \ref{FIG:1}). We utilize powerful MLLMs to synthesize the Chain-of-Thought (CoT) datasets QFSD-CoT and AgriInsect-CoT, which capture expert-informed pest feature reasoning trajectories. Building on these datasets, we propose \textbf{Pest-Thinker}, a two-stage knowledge-driven framework that enhances the visual reasoning and morphological comprehension of MLLMs. We first employ CoT trajectories to perform cold-start SFT, enabling the model to rapidly align with the desired output format and initialize the reasoning process for pest features. We subsequently apply the Group Relative Policy Optimization (GRPO) algorithm for RL. Furthermore, we innovatively introduce a feature reward that incentivizes description and reasoning over observable morphological traits, where the reasoning process is scored by the LLM-as-a-Judge method based on expert knowledge. Our method achieves superior performance across full-set, few-shot and cross-category generalization experimental settings, demonstrating its capacity to endow MLLMs with the ability to genuinely learn, understand, and reason about the essential characteristics of pests.

Our contributions are summarized as follows:
\begin{itemize}
    \item We propose Pest-Thinker, the first systematic exploration of knowledge-driven morphology learning and reasoning for MLLMs based on the RL paradigm in the context of agricultural pest analysis.
    \item We construct two high-quality pest benchmarks, QFSD and AgriInsect, annotated with expert-derived pest feature knowledge, enabling both SFT and RL training for MLLMs in pest-centric learning tasks.
    \item We introduce a fine-grained feature reward mechanism for RL fine-tuning, which effectively incentivizes the generalizable reasoning abilities of MLLMs toward pest visual features, achieving strong performance in both in-domain and out-of-domain morphological understanding.
    \item Extensive experiments conducted across multiple settings validate the effectiveness and robustness of Pest-Thinker, highlighting its practical potential for real-world pest monitoring and management applications.
\end{itemize}

%% file: sec/2_relatedwork.tex
\section{Related Work}
\label{sec:related}


\begin{figure*}[ht]
	\centering
        \includegraphics[width=0.93\linewidth]{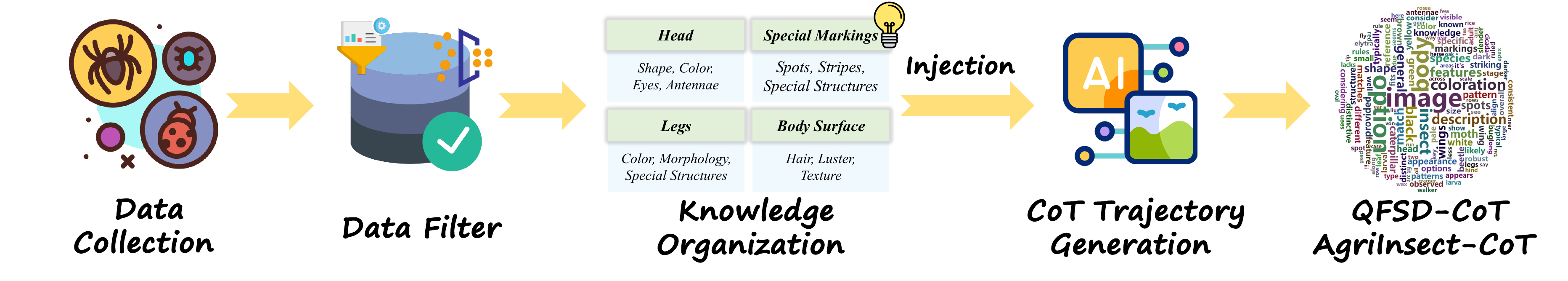}
	\caption{The data construction pipeline of QFSD-CoT and AgriInsect-CoT, which serves to construct CoT trajectories reflecting expert-informed reasoning about pest morphology.}
\label{FIG:data}
\end{figure*}
\begin{figure}[t]
	\centering
        \includegraphics[width=0.93\linewidth]{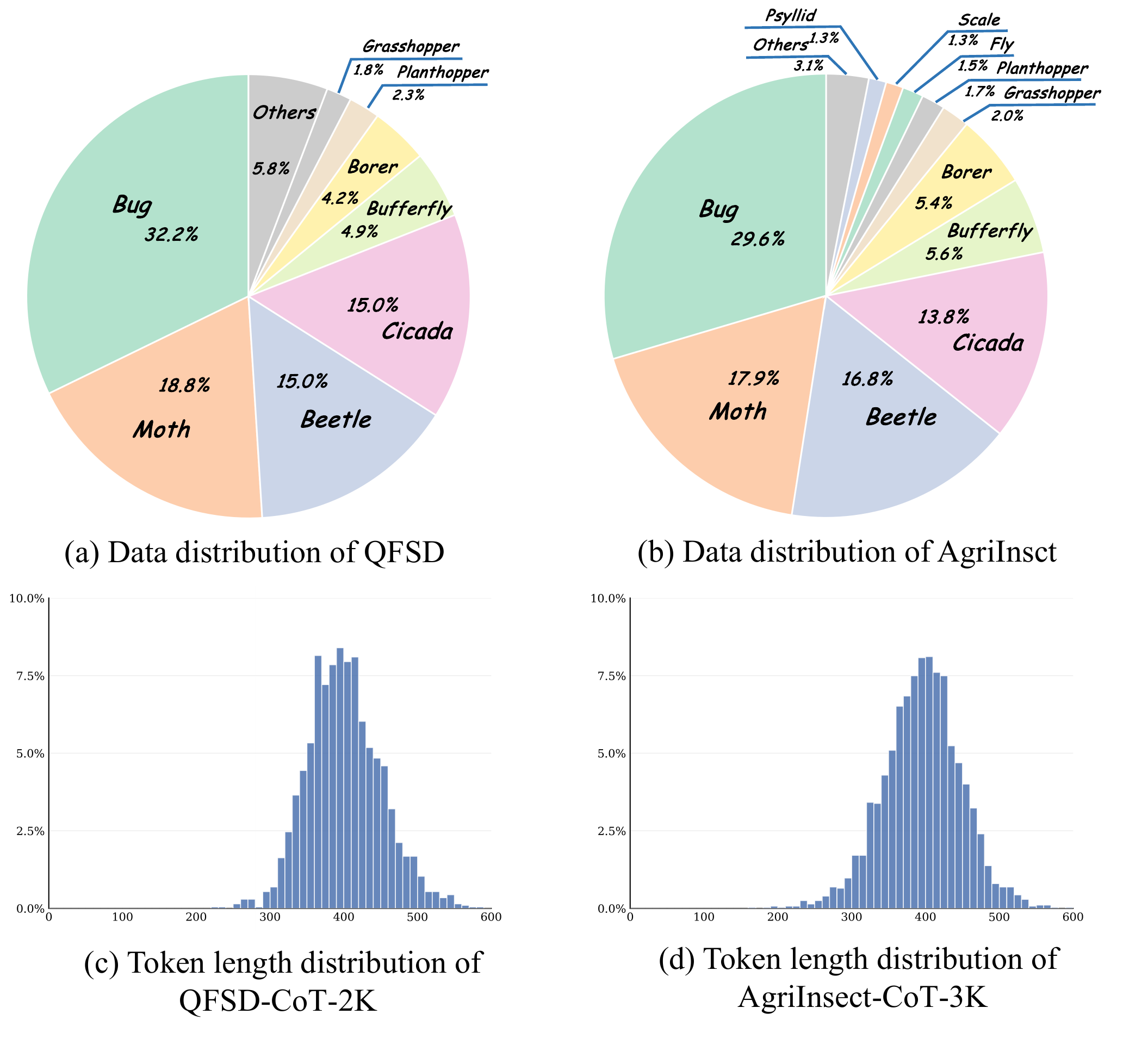}
	\caption{Data distribution of QFSD and AgriInsect datasets, along with the corresponding distribution of CoT token lengths.}
\label{FIG:dist}
\end{figure}
\subsection{Agricultural Pest Learning}
In recent years, deep learning has seen increasing applications in agricultural pest recognition. Most existing methods leverage backbone networks pretrained on general datasets and fine-tune them on pest-specific datasets~\cite{Hu2024, Butera2022}. More recent studies introduce architectures such as Mamba to design specialized networks~\cite{Saranya2024, Wang2025b}. This trend has recently evolved with the emergence of MLLMs~\cite{comanici2025gemini, bai2025qwen2, zhu2025internvl3}, which have demonstrated revolutionary performance across a wide array of visual understanding tasks. Inspired by recent successes, research interest in applying MLLMs to smart agriculture has surged. Representative efforts such as Insect-Foundation~\cite{truong2025insect} and AgriGPT-VL~\cite{yang2025agrigptvl, yang2025agrigpt} have developed agriculture-oriented MLLMs trained on large-scale datasets. However, these pioneering models rely heavily on data-driven SFT, leading to substantial practical constraints (e.g. large data requirements, high computational costs). Therefore, they remain poorly suited to real-world agricultural applications, which typically involve learning a limited number of common pest species.
\subsection{Multimodal Reasoning in MLLMs}

The application of reinforcement learning (RL) in post-training has been shown to yield substantial improvements in the reasoning capabilities of large models. DeepSeek-R1 \cite{guo2025deepseek} and OpenAI-o3 \cite{openai2025thinking} have demonstrated the potential of RL to enhance the consistency and depth of reasoning in complex tasks. In the visual domain, a growing body of work has extended the application of RL to MLLMs, employing RL to optimize vision-language alignment and multi-step reasoning ~\cite{wang2025videorft,qiu2025metis,guo2025ssl4rl}. VisionThink \cite{yang2025visionthink} emphasizes efficient visual reasoning under limited computational budgets. Vision-SR1 \cite{li2025self} applies reinforcement fine-tuning to optimize stepwise reasoning, particularly for fine-grained vision-text alignment. Vision-G1 \cite{zha2025vision} combines large-scale multimodal pretraining with RL-based reasoning to enhance generalization across diverse visual tasks. Vision-R1 \cite{huang2025visionr1} employs cold-start initialization and progressive RL training strategies to substantially improve multimodal mathematical reasoning. Collectively, these studies affirm the efficacy of RL in enhancing the visual reasoning capabilities of MLLMs. However, the practical application of these models in agricultural scenarios faces significant limitations, primarily attributable to data scarcity and their limited capacity for fine-grained feature understanding. In this work, we leverage RL to optimize key feature attention and multi-step reasoning, enabling precise identification of complex pest species. This approach concurrently enhances reasoning stability and interpretability in complex agricultural environments and offers robust support for precision pest monitoring and decision-making in smart agriculture.

%% file: sec/4_Method.tex
\section{Methods}

\begin{figure*}[ht]
	\centering
        \includegraphics[width=\linewidth]{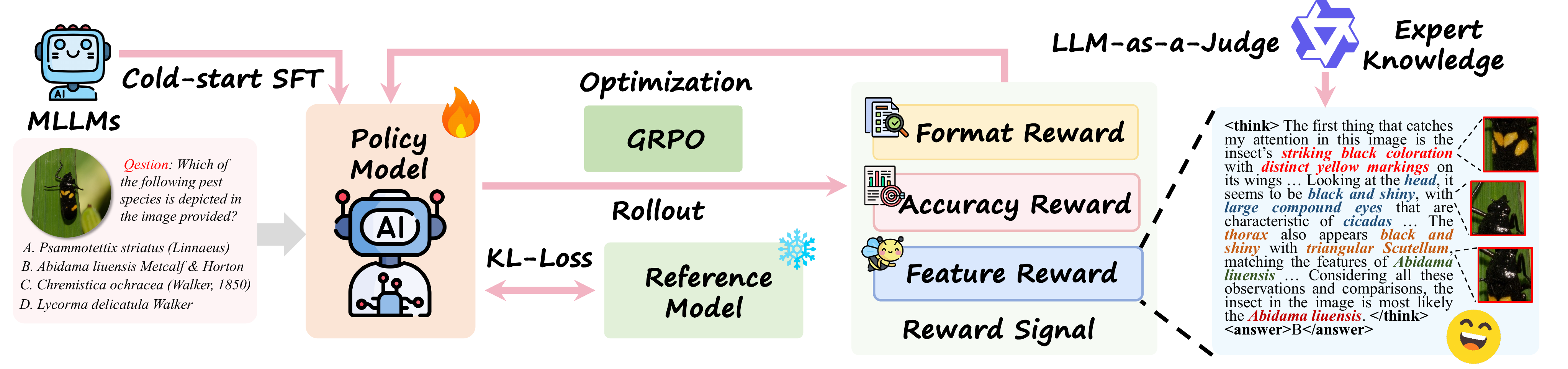}
	\caption{Illustration of the Pest-Thinker two-stage training paradigm. The feature reward signal encourages the model to think and reason about the pest morphological features, where different colors denote features corresponding to distinct body parts.}
\label{FIG:rl}
\end{figure*}
\begin{figure*}[ht]
	\centering
        \includegraphics[width=\linewidth]{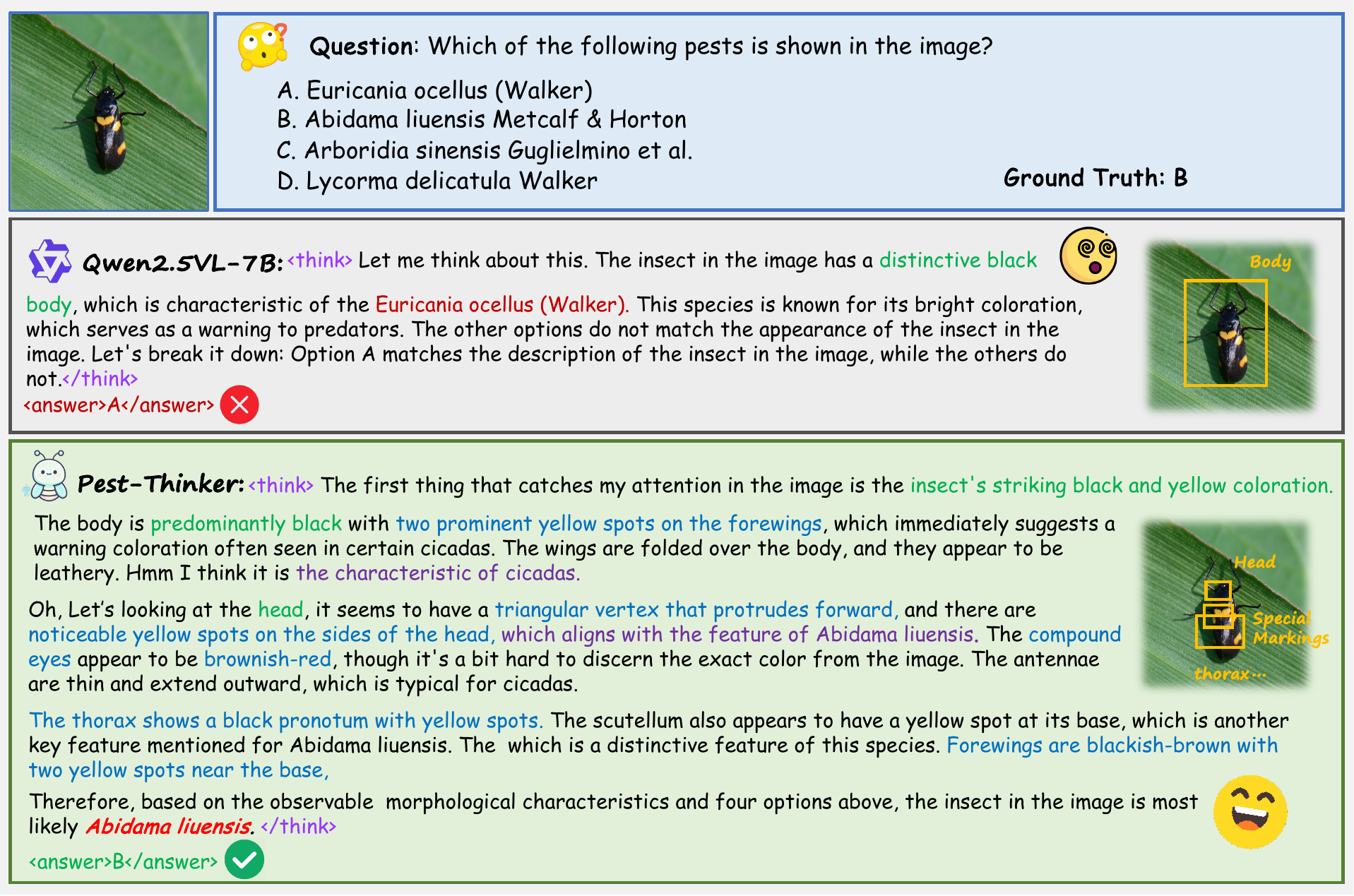}
	\caption{Comparison of reasoning traces generated by Qwen2.5VL-7B and Pest-Thinker on the QFSD dataset.}
\label{FIG:case_example}
\end{figure*}
\subsection{Dataset Construction}
\noindent \textbf{Data Collection and Curation.} To address the scarcity of high-quality pest training data and the inherently small size of individual pests, as well as to better simulate data distributions encountered in real-world agricultural environments, we captured and curated high-resolution pest images from multiple rural regions, ensuring clear visibility of their body surface characteristics. Our dataset encompasses a diverse range of major pest taxa, covering approximately thirteen general categories such as beetles, borers, cicadas, and moths. In contrast to existing pest benchmarks \cite{xie2015automatic, liu2022dataset, FarmInsect}, we adopt scientific latin name to classify pest species at a finer granularity, thereby encouraging the model to acquire a deeper comprehension of pest morphological knowledge. Following data filtering and refinement, we established two high-quality datasets, QFSD and AgriInsect, which contain 7,054 images of 141 species and 9,452 images of 200 species, respectively.

\noindent \textbf{Feature Knowledge Organization.} We commissioned experienced entomologists to annotate the key morphological features of each pest according to distinct body parts, categorizing them into seven regions: Head, Thorax, Abdomen, Wing, Leg, Body Surface, and Special Markings. For pests exhibiting multiple morphologies within the same species, we provided comprehensive annotations covering all observed variations, enabling the model to autonomously reason and learn from these diverse examples during training.

\noindent \textbf{CoT Trajectory Generation.} To effectively implement cold-start SFT, we employed Qwen2.5-VL-72B-Instruct \cite{bai2025qwen2} to generate CoT reasoning trajectories. During this process, we integrated our collected pest morphology knowledge into the trajectories, guiding the model to progressively attend to pest-specific characteristics and reason toward accurate answers. Using this approach, we constructed two high-quality cold-start datasets, QFSD-CoT-2K and AgriInsect-CoT-3K. The complete data construction pipeline is illustrated in Figure \ref{FIG:data}. We visualize the corresponding data distributions and the CoT token length distributions in Figure \ref{FIG:dist}. The remaining data are utilized to construct QFSD-RL-4K and AgriInsect-RL-6K, which served as training datasets for the reinforcement learning stage.

\subsection{Training Strategy}
\noindent \textbf{Cold-start SFT Fine-tuning.} To enable MLLMs to rapidly acquire agricultural scenario knowledge and enhance Pest-Thinker’s capability to generate structured reasoning trajectories, we first conduct cold-start Supervised Fine-Tuning (SFT) to initialize the model. During this process, the model’s reasoning steps are enclosed within {\textit{\textless think\textgreater\textless/think\textgreater}} tags, while the final predicted answer is placed within {\textit{\textless answer\textgreater\textless/answer\textgreater}} tags. The overall cold-start process can be formally defined as follows:
\begin{align}
\mathcal{L}_{\text{SFT}}(\theta) = -\mathbb{E}_{(\mathcal{Q}, \mathcal{Y}) \sim \mathcal{D}_{\text{SFT}}} \sum_{i=1}^{||\mathcal{Y}||} \log p_{\theta} \left( \mathcal{Y}_i \mid \mathcal{Q}, \mathcal{Y}_{<i} \right)
\end{align}
where $D_{\text{SFT}}$ is the cold-start dataset, $\mathcal{Q}$ denotes the user query, $\mathcal{Y}$ represents the corresponding target reasoning trajectory, and $p_{\theta}$ indicates the model parameterized by $\theta$.

\noindent \textbf{GRPO for Morphological Reasoning.} To enhance the model’s capability to reason about intricate pest morphological characteristics and generalize across varying developmental states, we employed the Group Relative Policy Optimization (GRPO) algorithm for fine-tuning in stage 2. GRPO generates $G$ candidate responses $\{o_1, o_2, ..., o_G\}$ for each query $\mathcal{Q}$. These responses are evaluated by the reward functions to produce corresponding reward values $\{r_1, r_2, …, r_G\}$, which are then normalized  and used to compute the advantage $A_i$ of each response:
\begin{align}
A_i=\frac{r_i-\text{mean}(\{r_1,r_2,\cdots,r_K\})}{\text{std}(\{r_1,r_2,\cdots,r_K\})}
\end{align}
where $\text{mean}(\cdot)$ and $\text{std}(\cdot)$ denote the mean and standard deviation of the rewards.  The policy model $\pi_{\theta}$ is subsequently optimized by maximizing following objective:
\begin{small}
\begin{align}
\mathcal{J}_{\mathrm{GRPO}}(\theta) = & \mathbb{E}_{(\mathcal{Q},\{o_{i}\})}\frac{1}{G}\sum_{i=1}^{G} \left[ \min \left(s_{\theta} \cdot A_{i}, \mathrm{clip_{1-\epsilon}^{1+\epsilon}}\left(s_{\theta} \right) \cdot A_{i} \right) \right. \nonumber \\
 & \left. -\beta\mathbb{D}_{\mathrm{KL}}(\pi_{\theta}||\pi_{\mathrm{ref}}) \right]
\end{align}
\end{small}
\noindent Here, $\pi_{\theta}$ and  $\pi_{\theta_{\mathrm{odd}}}$ denote the current and old policy models, respectively. The $s_{\theta} = \pi_{\theta} / \pi_{\theta_{\mathrm{odd}}}$ represents the policy ratio. $\pi_{\mathrm{ref}}$ is the reference policy, with $\beta$ serving as the KL-divergence regularization coefficient. The parameter $\epsilon$ defines a clipping range $(1-\epsilon, 1+\epsilon)$ on the policy ratio $s_{\theta}$, which prevents the policy update from deviating excessively from the reference model.

\noindent \textbf{Reward Modeling.}
During RL training, we employ three types of reward signals for joint optimization:
\begin{itemize}
    \item \textbf{Format Reward.} We adopt the widely-used binary format reward $R_{format}$ to constrain and regulate the model’s structured reasoning and answer generation. Specifically, the reasoning process is also constrained within {\textit{\textless think\textgreater\textless/think\textgreater}} tags, while answer is required to appear within {\textit{\textless answer\textgreater\textless/answer\textgreater}} tags. A reward value of 1 is assigned if the model output adheres to the prescribed format, and 0 otherwise.
    \item \textbf{Accuracy Reward.} This reward is designed to evaluate whether the model’s discrete answer matches the ground truth. The accuracy reward $R_{acc}$ is also binary in nature, assigning a value of 1 for a correct response and applying a penalty (0) for an incorrect one.
    \item \textbf{Feature Reward.} As illustrated in Figure \ref{FIG:rl}, the feature reward $R_{feature}$ is central to the Pest-Thinker. Its primary objective is to incentivize the model's reasoning on pest morphology, compelling it to analyze and differentiate key features in a manner akin to an entomologist. This mechanism rewards the model for explicitly describing observed morphological traits. The model's output are evaluated against expert-curated knowledge via the LLM-as-a-Judge approach. A score (0–10) is assigned and subsequently normalized to derive the final reward value. The detailed scoring prompt design is available in the supplementary materials.
\end{itemize}
In general, the overall reward for a single sample is computed as follows:
\begin{align}
R = R_{format} + R_{acc} + R_{feature}
\end{align}
After RL training, Pest-Thinker demonstrates a strong capability to understand and reason about the visual characteristics of pests, enabling multi-level analysis of morphological features across different body parts. Representative reasoning examples are shown in Figure \ref{FIG:case_example}.

%% file: sec/5_Experiment.tex
\section{Experiments}
\label{sec:experiments}

\begin{table*}[ht]
\centering
\caption{Quantitative comparison on the QFSD benchmark. The best results among open-source models are shown in \textbf{bold}, while the second-best results are \underline{underlined}. All evaluation metrics are reported as percentages (\%), where higher values indicate better performance.}
\small
\renewcommand{\arraystretch}{0.85}
\resizebox{\textwidth}{!}{%
\begin{tabular}{l*{9}{c}c}
\toprule
 & \multicolumn{9}{c}{\textbf{General Pest Categories}} &  \\
\cmidrule(lr){2-10}
\textbf{Model} & \textbf{Beetle} & \textbf{Borer} & \textbf{Bug} & \textbf{Butterfly} & \textbf{Cicada} & \textbf{Grasshopper} & \textbf{Moth} & \textbf{Planthopper} & \textbf{Scale} & \textbf{Overall} \\
\midrule
\rowcolor{myblue}\multicolumn{11}{c}{\textbf{\textit{Proprietary Models}}} \\ 
\midrule
GPT-5 & 85.44 & 92.59 & 89.32 & 96.67 & 62.50 & 54.55 & 87.29 & 64.71 & 87.50 & 83.28 \\
GPT-5-mini & 78.64 & 74.07 & 73.79 & 90.00 & 51.04 & 36.36 & 83.05 & 47.06 & 50.00 & 72.56 \\
Gemini-2.5-Flash & 83.95 & 84.21 & 87.82 & 96.15 & 78.08 & 70.00 & 89.16 & 75.00 & 40.00 & 84.94 \\
\midrule
\rowcolor{mygreen}\multicolumn{11}{c}{\textbf{\textit{Open-source Vanilla Models}}} \\ 
\midrule
InternVL3-8B & 42.72 & 48.15 & 40.78 & 53.33 & 46.88 & 0.00 & 49.15 & 29.41 & 0.00 & 42.74 \\
InternVL3.5-8B & 36.89 & 25.93 & 39.32 & 40.00 & 36.36 & 54.55 & 39.83 & 52.94 & 25.00 & 38.96 \\
Qwen2.5VL-3B-Instruct & 45.63 & 25.93 & 35.44 & 76.67 & 45.83 & 9.09 & 56.78 & 41.18 & 25.00 & 43.85 \\
Qwen2.5VL-7B-Instruct & 48.54 & 25.93 & 48.54 & 76.67 & 44.79 & 27.27 & 59.32 & 29.41 & 0.00 & 49.21 \\
Qwen2.5VL-32B-Instruct & 54.37 & 48.15 & 43.69 & \underline{86.67} & 50.00 & 54.55 & 62.71 & 70.59 & 25.00 & 52.68 \\
Qwen2.5VL-72B-Instruct & 71.84 & 51.85 & 60.68 & 73.33 & 51.04 & 54.55 & 66.95 & 52.94 & 62.50 & 62.46 \\
Qwen3VL-8B-Instruct & 47.52 & 43.48 & 34.95 & 70.37 & 46.81 & 27.27 & 55.26 & 40.00 & 62.50 & 44.89 \\
Qwen3VL-30B-Instruct & 54.00 & 65.38 & 51.96 & 80.00 & 45.74 & 18.18 & 70.34 & \underline{76.47} & \underline{75.00} & 57.03 \\
\midrule
\rowcolor{myorange}\multicolumn{11}{c}{\textbf{\textit{Open-source Reasoning Models}}} \\ 
\midrule
Qwen3VL-8B-Thinking & 61.62 & 37.50 & 51.87 & 5.00 & 41.94 & 63.64 & 63.21 & 46.15 & 42.86 & 54.91 \\
Qwen3VL-30B-Thinking & 61.29 & 56.00 & 63.21 & 77.78 & 57.14 & 22.22 & 73.45 & 71.43 & 25.00 & 63.39 \\
VisionThink-Efficient-7B & 57.28 & 37.04 & 57.28 & 76.67 & 46.88 & 54.55 & 65.25 & 41.18 & 12.50 & 56.31 \\
VisionThink-General-7B & 55.34 & 51.85 & 55.83 & 83.33 & 44.79 & 54.55 & 66.95 & 47.06 & 0.00 & 56.15 \\
Vision-SR1-7B & 47.57 & 29.63 & 47.57 & 66.67 & 44.79 & 27.27 & 42.37 & 47.06 & 12.50 & 46.06 \\
Vision-G1-7B & 55.34 & 48.15 & 54.85 & 80.00 & 47.92 & 45.45 & 59.32 & 52.94 & 0.00 & 54.42 \\
\midrule
\rowcolor{myred}\multicolumn{11}{c}{\textbf{\textit{Our Models}}} \\ 
\midrule
Pest-Thinker-SFT-3B (ours) & 71.84 & 62.96 & 65.73 & 80.00 & 65.62 & 72.73 & 66.27 & 47.06 & 62.50 & 63.71 \\
Pest-Thinker-SFT-7B (ours) & 72.52 & 74.07 & 77.86 & 83.33 & \underline{78.54} & 74.62 & 77.29 & 41.18 & \underline{75.00} & 75.33 \\
Pest-Thinker-3B (ours) & \underline{76.55} & \underline{82.59} & \textbf{88.83} & {80.00} & 73.33 & \underline{81.82} & \underline{78.14} & 52.94 & 62.50 & \underline{75.96} \\
Pest-Thinker-7B (ours) & \textbf{81.26} & \textbf{92.59} & \underline{86.12} & \textbf{96.67} & \textbf{87.75} & \textbf{90.91} & \textbf{84.37} & \textbf{84.12} & \textbf{87.50} & \textbf{84.32} \\
\bottomrule
\end{tabular}%
}
\label{tab:qfsd}
\end{table*}

\subsection{Experimental Setup}
\noindent \textbf{Datasets and Benchmarks.} All experiments are conducted on the QFSD and AgriInsect benchmarks. For the full-set overall experiments, approximately 15\% of the data from each benchmark is reserved to construct the test set. For the few-shot experiments, the models were trained under four few-shot scenarios: 1-shot, 2-shot, 4-shot, and 8-shot. For the few-shot experiments, distinct N-shot samples are employed in both the SFT and RL stages. The evaluation was performed on the same full-set test set to maintain strict consistency and enable direct comparability across varying degrees of data scarcity. For the generalization experiments, we select 24 intra-species pest classes exhibiting diverse visual morphologies (e.g., instar stages, different growth states) and assign a single sub-category as the training set, while using the remaining sub-categories as the test set. The categories in the training and test sets are mutually exclusive, ensuring a rigorous evaluation of the model’s  generalization capability. 

\noindent \textbf{Baseline Models.} To comprehensively evaluate the effectiveness of Pest-Thinker, we perform extensive comparisons against three categories of baseline MLLMs: (i) commercial proprietary models, including GPT-5 \cite{openai2025gpt5}, GPT-5-mini \cite{openai2025gpt5} and Gemini-2.5-Flash \cite{comanici2025gemini}; (ii) open-source vanilla models, including InternVL3-8B \cite{zhu2025internvl3},  InternVL3.5-8B \cite{wang2025internvl3}, Qwen2.5VL-Instruct series \cite{bai2025qwen2} and Qwen3VL-Instruct series \cite{yang2025qwen3}; (iii) open-source reasoning models, comprising Qwen3VL-Thinking series \cite{yang2025qwen3}, VisionThink-Efficient-7B \cite{yang2025visionthink}, VisionThink-General-7B \cite{yang2025visionthink}, Vision-SR1-7B \cite{li2025self}, and Vision-G1-7B \cite{zha2025vision}.

\noindent \textbf{Implement Details.} We train our model using 8 NVIDIA H100 GPUs. The Qwen2.5VL-3B-Instruct and Qwen2.5VL-7B-Instruct \cite{bai2025qwen2} serve as the base models for all training stages. For the overall experiments, during the SFT phase, the models are trained on the QFSD-CoT-2K and AgriInsect-CoT-3K datasets for 1 epoch, with a learning rate of $1 \times 10^{-5}$ and a batch size of 1. In the subsequent RL fine-tuning stage, we utilize the QFSD-RL-4K and AgriInsect-RL-6K datasets. The initial learning rate for the RL stage is set to $1 \times 10^{-6}$ with a batch size of 1. To maintain training stability, we apply a weight decay of 0.01 and clip the maximum gradient norm to 5. Both training stages adopt a consistent prompting template, as detailed in the supplementary material. For computational efficiency, all input images are uniformly resized to a resolution of $512 \times 512$. We employ Qwen3-235B-A22B-Instruct as the feature reward scoring model.

\begin{table*}[ht]
\centering
\caption{Quantitative comparison on the AgriInsect benchmark. The best results among open-source models are shown in \textbf{bold}, while the second-best results are \underline{underlined}. All evaluation metrics are reported as percentages (\%), where higher values indicate better performance.}
\small
\renewcommand{\arraystretch}{0.85}
\resizebox{\textwidth}{!}{%
\begin{tabular}{l*{9}{c}c}
\toprule
 & \multicolumn{9}{c}{\textbf{General Pest Categories}} &  \\
\cmidrule(lr){2-10}
\textbf{Model} & \textbf{Beetle} & \textbf{Borer} & \textbf{Bug} & \textbf{Butterfly} & \textbf{Cicada} & \textbf{Grasshopper} & \textbf{Moth} & \textbf{Planthopper} & \textbf{Scale} & \textbf{Overall} \\
\midrule
\rowcolor{myblue}\multicolumn{11}{c}{\textbf{\textit{Proprietary Models}}} \\ 
\midrule
GPT-5 & 87.42 & 84.00 & 83.59 & 100.00 & 81.03 & 64.71 & 85.62 & 44.44 & 54.55 & 83.24 \\
GPT-5-mini & 82.12 & 72.00 & 69.92 & 85.11 & 68.97 & 47.06 & 78.43 & 50.00 & 54.55 & 73.34 \\
Gemini-2.5-Flash & 91.24 & 77.08 & 84.94 & 93.02 & 70.91 & 50.00 & 87.14 & 61.11 & 60.00 & 82.05 \\
\midrule
\rowcolor{mygreen}\multicolumn{11}{c}{\textbf{\textit{Open-source Vanilla Models}}} \\ 
\midrule
InternVL3-8B & 41.06 & 48.00 & 39.84 & 51.06 & 50.00 & 11.76 & 46.41 & 22.22 & 27.27 & 42.03 \\
InternVL3.5-8B & 37.75 & 30.00 & 37.50 & 42.55 & 32.76 & 35.29 & 35.95 & 11.11 & 18.18 & 35.51 \\
Qwen2.5-VL-3B-Instruct & 45.70 & 32.00 & 39.84 & 63.83 & 40.52 & 11.76 & 43.79 & 38.89 & 18.18 & 42.14 \\
Qwen2.5VL-7B-Instruct & 51.66 & 34.00 & 51.95 & 57.45 & 43.97 & 17.65 & 55.56 & 33.33 & 54.55 & 49.48 \\
Qwen2.5VL-32B-Instruct & 60.93 & 54.00 & 46.88 & 72.34 & 38.79 & 41.18 & 66.01 & 50.00 & 27.27 & 52.62 \\
Qwen2.5VL-72B-Instruct & 64.67 & 44.68 & 56.47 & 82.98 & 50.86 & 41.18 & 64.43 & 55.56 & 54.55 & 59.06 \\
Qwen3VL-8B-Instruct & 51.35 & 42.55 & 40.71 & 77.78 & 39.13 & 17.65 & 44.22 & 50.00 & 33.33 & 44.68 \\
Qwen3VL-30B-Instruct & 56.08 & \textbf{68.00} & 53.54 & 85.11 & 49.14 & 35.29 & 64.24 & 55.56 & 45.45 & 56.81 \\
\midrule
\rowcolor{myorange}\multicolumn{11}{c}{\textbf{\textit{Open-source Reasoning Models}}} \\ 
\midrule
Qwen3VL-8B-Thinking & 59.59 & 47.62 & 46.09 & 60.87 & 43.64 & 35.29 & 55.32 & 47.06 & {55.56} & 50.51 \\
Qwen3VL-30B-Thinking & 67.67 & 53.49 & 61.44 & 74.42 & 51.38 & 23.08 & 70.83 & 57.14 & \textbf{85.71} & 61.97 \\
VisionThink-Efficient-7B & 58.94 & 44.00 & 50.78 & 68.09 & 49.14 & 17.65 & 64.05 & 55.56 & 27.27 & 53.78 \\
VisionThink-General-7B & 58.28 & 40.00 & 50.78 & 74.47 & 52.59 & 35.29 & 64.71 & 44.44 & 36.36 & 54.60 \\
Vision-SR1-7B & 56.29 & 46.00 & 48.05 & 68.09 & 48.28 & 41.18 & 59.48 & 55.56 & 36.36 & 51.92 \\
Vision-G1-7B & 54.97 & 38.00 & 50.78 & 68.09 & 53.45 & 29.41 & 60.13 & 44.44 & 36.36 & 52.74 \\
\midrule
\rowcolor{myred}\multicolumn{11}{c}{\textbf{\textit{Our Models}}} \\ 
\midrule
Pest-Thinker-SFT-3B (ours) & 60.26 & 62.00 & 66.80 & 77.23 & 61.21 & 41.18 & 67.97 & 55.56 & 18.18 & 64.14 \\
Pest-Thinker-SFT-7B (ours) & \underline{76.75} & 60.00 & 70.31 & 76.77 & \underline{79.31} & \underline{69.31} & 74.12 & \underline{78.89} & 45.45 & 73.00 \\
Pest-Thinker-3B (ours) & 69.54 & 62.00 &  \underline{77.73} &  \underline{85.11} & 77.59 & 58.82 & \underline{79.74} & 55.56 & \underline{61.11} & \underline{73.92} \\
Pest-Thinker-7B (ours) & \textbf{84.77} & \underline{66.00} & \textbf{86.33} & \textbf{95.74} & \textbf{81.90} & \textbf{70.59} & \textbf{87.58} & \textbf{88.89} & 54.55 & \textbf{82.77} \\
\bottomrule
\end{tabular}%
}
\label{tab:agriinsect}
\end{table*}


\begin{table}[htbp]
    \centering
    \caption{Few-shot quantitative comparison on QFSD. All evaluation metrics are reported as percentages (\%). Here, $\Delta$ represents the performance difference between the RL and SFT stages.}
    \footnotesize
    \renewcommand{\arraystretch}{0.75}
    \resizebox{0.48\textwidth}{!}{%
    \begin{tabular}{c*{6}{c}}
    \toprule
     & \multicolumn{5}{c}{\textbf{General Pest Categories}} &  \\
    \cmidrule(lr){2-6}
    \textbf{Model} & \textbf{Beetle} & \textbf{Bug} & \textbf{Butterfly} & \textbf{Cicada} & \textbf{Moth} & \textbf{Overall} \\
    \midrule
    \multicolumn{7}{c}{\textbf{\textit{Qwen2.5VL-3B}}} \\
    \midrule
    Baseline & 45.63 & 35.44 & 76.67 & 45.83 & 56.78 & 43.85 \\
    \midrule
    \multicolumn{7}{c}{\textbf{\textit{1-shot}}} \\
    \cmidrule(lr){1-7}
    + SFT & 46.60 & 42.72 & 60.00 & 42.71 & 47.46 & 44.32 \\
    + RL & 51.46 & 54.15 & 76.67 & 47.92 & 64.41 & 54.98 \\
    \rowcolor{mycream}$\Delta$ & 4.86 & 11.43 & 16.67 & 5.21 & 16.95 & 10.66 \\
    \midrule
    \multicolumn{7}{c}{\textbf{\textit{2-shot}}} \\
    \cmidrule(lr){1-7}
    + SFT & 48.54 & 49.03 & 63.33 & 47.92 & 51.39 & 48.10 \\
    + RL & 51.46 & 60.00 & 66.67 & 50.00 & 63.56 & 56.96 \\
    \rowcolor{mycream}$\Delta$ & 2.92 & 10.97 & 3.34 & 2.08 & 12.17 & 8.86 \\
    \midrule
    \multicolumn{7}{c}{\textbf{\textit{4-shot}}} \\
    \cmidrule(lr){1-7}
    + SFT & 50.49 & 63.59 & 70.00 & 49.47 & 61.02 & 54.57 \\
    + RL & 54.36 & 64.08 & 83.33 & 54.17 & 64.41 & 59.67 \\
    \rowcolor{mycream}$\Delta$ & 3.87 & 0.49 & 13.33 & 4.70 & 3.39 & 5.10 \\
    \midrule
    \multicolumn{7}{c}{\textbf{\textit{8-shot}}} \\
    \cmidrule(lr){1-7}
    + SFT & 53.40 & 58.25 & 73.33 & 47.92 & 57.62 & 58.04 \\
    + RL & 56.34 & 64.56 & 80.00 & 58.33 & 68.64 & 62.78 \\
    \rowcolor{mycream}$\Delta$ & 2.94 & 6.31 & 6.67 & 10.41 & 8.47 & 4.74 \\
    \bottomrule
    \end{tabular}%
    }
    \label{tab:fewshot1}
    \end{table}

\begin{table}[htbp]
    \centering
    \caption{Few-shot quantitative comparison on AgriInsect. All evaluation metrics are reported as percentages (\%). Here, $\Delta$ represents the performance difference between the RL and SFT stages.}
    \footnotesize
    \renewcommand{\arraystretch}{0.75}
    \resizebox{0.48\textwidth}{!}{%
    \begin{tabular}{c*{6}{c}}
    \toprule
     & \multicolumn{5}{c}{\textbf{General Pest Categories}} &  \\
    \cmidrule(lr){2-6}
    \textbf{Model} & \textbf{Beetle} & \textbf{Bug} & \textbf{Butterfly} & \textbf{Cicada} & \textbf{Moth} & \textbf{Overall} \\
    \midrule
    \multicolumn{7}{c}{\textbf{\textit{Qwen2.5VL-3B}}} \\
    \midrule
    Baseline & 45.70 & 39.84 & 63.83 & 40.52 & 43.79 & 42.14 \\
    \midrule
    \multicolumn{7}{c}{\textbf{\textit{1-shot}}} \\
    \cmidrule(lr){1-7}
    + SFT & 43.71 & 37.89 & 65.96 & 39.32 & 45.10 & 40.86 \\
    + RL & 49.67 & 52.36 & 68.09 & 54.31 & 58.85 & 52.04 \\
    \rowcolor{mycream}$\Delta$ & 5.96 & 14.47 & 2.13 & 14.99 & 13.75 & 11.18 \\
    \midrule
    \multicolumn{7}{c}{\textbf{\textit{2-shot}}} \\
    \cmidrule(lr){1-7}
    + SFT & 46.36 & 46.87 & 70.21 & 50.00 & 50.33 & 48.89 \\
    + RL & 50.99 & 52.73 & 76.60 & 53.45 & 57.51 & 54.50 \\
    \rowcolor{mycream}$\Delta$ & 4.63 & 7.18 & 6.39 & 3.45 & 5.23 & 5.61 \\
    \midrule
    \multicolumn{7}{c}{\textbf{\textit{4-shot}}} \\
    \cmidrule(lr){1-7}
    + SFT & 52.98 & 53.91 & 68.09 & 55.17 & 60.13 & 53.95 \\
    + RL & 56.95 & 58.98 & 78.72 & 57.76 & 65.36 & 58.40 \\
    \rowcolor{mycream}$\Delta$ & 3.97 & 5.07 & 10.63 & 2.59 & 5.23 & 4.45 \\
    \midrule
    \multicolumn{7}{c}{\textbf{\textit{8-shot}}} \\
    \cmidrule(lr){1-7}
    + SFT & 56.95 & 53.52 & 70.21 & 58.62 & 64.05 & 56.58 \\
    + RL & 61.58 & 58.20 & 80.85  & 64.66 & 71.90 & 60.85 \\
    \rowcolor{mycream}$\Delta$ & 4.63 & 4.68 & 10.64 & 6.04 & 7.85 & 4.27 \\
    \bottomrule
    \end{tabular}%
    }
    \label{tab:fewshot2}
    \end{table}

\begin{table*}[ht]
    \centering
    \caption{Experimental results comparing different knowledge guidance and reward modeling mechanism across QFSD and AgriInsect datasets. The table shows performance metrics for four general pest categories  and overall accuracy. All ablation experiments were conducted using the Qwen2.5VL-3B-Instruct model.} 
    \small
    \setlength{\tabcolsep}{4pt}
    \resizebox{\textwidth}{!}{
    \begin{tabular}{lcccccc cccccc}
    \toprule
    & \multicolumn{6}{c}{\textbf{QFSD}} & \multicolumn{6}{c}{\textbf{AgriInsect}} \\
    \cmidrule(lr){2-7} \cmidrule(lr){8-13}
    \textbf{Model} & \textbf{Beetle} & \textbf{Bug} & \textbf{Butterfly} & \textbf{Cicada} & \textbf{Moth} & \textbf{Overall} & \textbf{Beetle} & \textbf{Bug} & \textbf{Butterfly} & \textbf{Cicada} & \textbf{Moth} & \textbf{Overall} \\
    \midrule
    \multicolumn{13}{l}{\textbf{Knowledge Effect}} \\
    RL only & 47.05 & 35.92 & 73.33 & 48.96 & 59.48 & 50.63 & 47.68 & 43.92 & 68.08 & 47.41 & 48.68 & 53.26 \\
    SFT + RL & 54.06 & 44.17 & 73.33 & 55.20 & 61.02 & 56.91 & 51.65 & 50.78 & 74.47 & 52.59 & 54.90 & 57.60 \\
    \midrule
    \multicolumn{13}{l}{\textbf{Reward Mechanism}} \\
    $R$ = $R_{format}$ + $R_{acc}$ & 70.87 & 62.56 & 76.68 & 64.58 & 68.97 & 67.64 & 59.60 &  68.50 & 80.85 & 62.93 & 66.01 & 66.24 \\
    $R$ = $R_{format}$ + $R_{acc}$ + $R_{feature}$ & \textbf{76.55} & \textbf{88.83} & \textbf{80.00} & \textbf{73.33} & \textbf{78.14} & \textbf{75.96} & \textbf{69.54} & \textbf{77.73} & \textbf{85.11} & \textbf{77.59} & \textbf{79.74} & \textbf{73.92} \\
    \bottomrule
    \end{tabular}
    }
\label{tab:ablation}
\end{table*}

\begin{table}[ht]
    \centering
    \caption{Intra-species generalization evaluation on AgriInsect. All evaluation metrics are reported as percentages (\%).}
    
    \small
    \renewcommand{\arraystretch}{1.1}
    \setlength{\tabcolsep}{6pt}
    \resizebox{0.48\textwidth}{!}{%
    \begin{tabular}{l*{5}{c}}
    \toprule
     & \multicolumn{4}{c}{\textbf{General Pest Categories}} &  \\
    \cmidrule(lr){2-5}
    \textbf{Model}  & \textbf{Borer} & \textbf{Bug} & \textbf{Cicada} & \textbf{Moth} & \textbf{Overall} \\
    \midrule
    \multicolumn{6}{c}{\textit{\textbf{Baseline Models}}} \\
    \cmidrule(lr){1-6}
    Qwen2.5VL-3B & 25.93 & 40.40 & 42.37 & 37.50 & 41.14 \\
    Qwen2.5VL-7B & 48.15 & 49.67 & 76.97 & 42.19 & 52.86 \\
    \midrule
    \multicolumn{6}{c}{\textit{\textbf{Our Models}}} \\
    \cmidrule(lr){1-6}
    Pest-Thinker-SFT-3B & 30.77 & 44.59 & 51.79 & 38.71 & 43.96 \\
    Pest-Thinker-SFT-7B & 52.38 & 50.39 & 75.47 & 45.65 & 53.68 \\
    Pest-Thinker-3B & 37.50 & 56.12 & 74.55 & 44.44 & 54.43 \\
    Pest-Thinker-7B & \textbf{70.83} & \textbf{69.06} & \textbf{87.27} & \textbf{63.63} & \textbf{70.55} \\
    \bottomrule
    \end{tabular}%
    }
    \label{tab:generalization}
    \end{table}

\subsection{Main Results.}
\noindent \textbf{Overall Experiments.} We performed comprehensive experiments on the QFSD and AgriInsect datasets. The results in Tables \ref{tab:qfsd} and \ref{tab:agriinsect} report the accuracy for a selection of nine representative general pest categories, alongside the overall performance. Our model improved substantially post-SFT, achieving accuracy comparable to Qwen2.5VL-72B-Instruct. Subsequent RL training delivered an additional 10\% improvement in overall performance, consistently exceeding open-source models across all categories. After RL optimization, the 7B model achieves performance comparable to advanced proprietary counterparts. This underscores our approach's efficacy in pest feature modeling, yielding robust morphological reasoning capabilities.

\noindent \textbf{Few-shot Experiments.} To evaluate the model's efficacy under data-scarce scenarios, we performed two few-shot experiments (Tables \ref{tab:fewshot1} and \ref{tab:fewshot2}). The results reveal that in these few-shot settings, the performance gains from SFT are marginal, occasionally resulting in performance declines. Conversely, RL yields substantial performance enhancements. This indicates that Pest-Thinker can rapidly acquire pest morphological knowledge and execute effective reasoning with minimal data, highlighting its robust suitability for real-world agricultural applications.

\noindent \textbf{Generalization Experiments.} To further evaluate the generalization reasoning capabilities of Pest-Thinker, we conducted the generalization experiment, the results of which are presented in Table \ref{tab:generalization}. This experiment employed 24 intra-specific pest categories exhibiting distinct visual states (e.g., different developmental stages, growth conditions, or morphological variations). These categories were selected to ensure complete mutual exclusivity between the training and test sets. The results demonstrate that Pest-Thinker effectively leverages morphological knowledge to identify pest morphologies unseen during training. This capability represents a substantial improvement over the baseline models, highlighting the robust generalization ability imparted by our proposed morphological knowledge reasoning trajectory and feature reward optimization.




\subsection{Ablation Experiments}
We performed ablation studies on the knowledge effect and reward mechanism of Pest-Thinker using the QFSD and AgriInsect datasets based on the Qwen2.5VL-3B-Instruct model, as summarized in Table \ref{tab:ablation}. 

\noindent \textbf{Knowledge Effect.} To evaluate the impact  of the constructed morphological knowledge, we conducted the ablation study in which the knowledge data was removed. The results reveal that without knowledge-driven learning, the model exhibits only marginal performance gains following post-training. This indicates a failure to acquire a genuine understanding of pest morphological features and underscores the significance of critical knowledge in guiding the model's comprehension.

\noindent \textbf{Reward Mechanism.} We investigated the effects of using only the format and accuracy rewards ($R$ = $R_{format}$ + $R_{acc}$), as well as the impact of incorporating the feature reward ($R$ = $R_{format}$ + $R_{acc}$ + $R_{feature}$), by evaluating two variations of the reward signals. This type of ablation was performed on the cold-start model. As presented in Table \ref{tab:ablation}, the inclusion of the feature reward leads to a considerable performance gain across all pest categories as well as the overall  performance, which highlights its essential contribution to the training process.




%% file: sec/6_Conclusion.tex
\section{Conclusion}
\label{sec:conclusion}
In this work, we propose Pest-Thinker, a novel two-stage knowledge-driven framework utilizing Reinforcement Learning (RL) to enhance the fine-grained visual reasoning of MLLMs for pest morphological understanding and learning. We first constructed two expert-annotated benchmarks, QFSD and AgriInsect, to synthesize the Chain-of-Thought (CoT) trajectory for cold-start SFT. We then applied GRPO algorithm guided by a novel fine-grained feature reward, scored by the LLM-as-a-Judge method, to encourage the model to reason over subtle morphological traits. Extensive experiments under full-set, data-scarce, and cross-category scenarios validate Pest-Thinker's effectiveness and robustness. This work presents the first systematic exploration of RL for fine-grained pest analysis, demonstrating a promising path for applying MLLMs to expert-level visual reasoning tasks in smart agriculture.
